# Statistical Limits of Convex Relaxations

Zhaoran Wang[*]   Quanquan Gu[†]   Han Liu[*]

**Abstract**

Many high dimensional sparse learning problems are formulated as nonconvex optimization. A popular approach to solve these nonconvex optimization problems is through convex relaxations such as linear and semidefinite programming. In this paper, we study the statistical limits of convex relaxations. Particularly, we consider two problems: Mean estimation for sparse principal submatrix and edge probability estimation for stochastic block model. We exploit the sum-of-squares relaxation hierarchy to sharply characterize the limits of a broad class of convex relaxations. Our result shows statistical optimality needs to be compromised for achieving computational tractability using convex relaxations. Compared with existing results on computational lower bounds for statistical problems, which consider general polynomial-time algorithms and rely on computational hardness hypotheses on problems like planted clique detection, our theory focuses on a broad class of convex relaxations and does not rely on unproven hypotheses.

## 1 Introduction

A broad variety of high dimensional statistical problems are formulated as nonconvex optimization. For example, sparse estimation can be formulated as optimization under $\ell_0$-norm constraints, where the $\ell_0$-norm is a pseudo-norm defined as the number of nonzero elements in a vector. To solve these nonconvex optimization problems, a popular approach is to resort to convex relaxations. Particularly, for sparse estimation, significant progress has been made by using $\ell_1$-norm as a convex relaxation for the nonconvex $\ell_0$-norm (see, e.g., Bühlmann and van de Geer (2011); Chandrasekaran et al. (2012) and the references therein).

In this paper, we study the statistical limits of convex relaxations. In particular, we focus on the sum-of-squares (SoS) hierarchy of convex relaxations (Lasserre, 2001; Parrilo, 2000, 2003), which is made up of a sequence of increasingly tighter convex relaxations based on semidefinite programming. We study the SoS hierarchy because it attains tighter approximations than other hierarchies such as the hierarchies proposed by Sherali and Adams (1990) and Lovász and Schrijver (1991), as well as their extensions (see Laurent (2003) for a comparison). Hence, the estimators in the SoS hierarchy achieve superior statistical performance than the estimators within other weaker hierarchies, which suggests the statistical limits of the SoS hierarchy are also the limits of weaker hierarchies.

[*]Department of Operations Research and Financial Engineering, Princeton University, Princeton, NJ 08544, USA; Email: {zhaoran, hanliu}@princeton.edu

[†]Department of Systems and Information Engineering, University of Virginia, Charlottesville, VA 22904, USA; e-mail: qg5w@virginia.edu

To demonstrate the statistical limits of convex relaxations, we focus on the examples of sparse principal submatrix estimation and stochastic block model estimation. In detail, for sparse principal submatrix estimation, we assume there is a $s^* \times s^*$ submatrix with elevated mean $\beta^*$ on the diagonal of a $d \times d$ noisy symmetric matrix. For stochastic block model estimation, we assume there exists a dense subgraph with $s^*$ nodes planted in an Erdős-Rényi graph with $d$ nodes. We denote by $\beta^*$ the edge probability of the subgraph. For both examples, our goal is to estimate $\beta^*$ under a challenging regime where $s^* = o\big[(d/\sqrt{\log d})^{2/3}\big]$ and $\log d = o(s^*)$. We prove the following information-theoretic lower bound

$$\inf_{\widehat{\beta}} \sup_{\mathbb{P} \in \mathcal{P}(s^*, d)} \mathbb{E}_{\mathbb{P}} \big|\widehat{\beta} - \beta^*\big| \geq C\sqrt{1/s^* \cdot \log(d/s^*)}, \tag{1.1}$$

where $\widehat{\beta}$ denotes any estimator, $\mathcal{P}(s^*, d)$ is the distribution family to be specified later and $C$ is an absolute constant. We prove that a computational intractable estimator $\widehat{\beta}^{\text{scan}}$ (to be specified later) attains the lower bound in (1.1). In order to achieve computational tractability, we consider convex relaxations of $\widehat{\beta}^{\text{scan}}$ that fall within the SoS and weaker hierarchies, which are denoted by $\mathcal{H}$. Let $C'$ be a positive absolute constant. We prove that under certain conditions,

$$\inf_{\widehat{\beta} \in \mathcal{H}} \sup_{\mathbb{P} \in \mathcal{P}(s^*, d)} \mathbb{E}_{\mathbb{P}} \big|\widehat{\beta} - \beta^*\big| \geq C'. \tag{1.2}$$

Together with (1.1), (1.2) illustrates the statistical limitations of a broad class of convex relaxations. Ignoring the logarithmic factor, (1.1) and (1.2) suggest there exists a gap of $\sqrt{s^*}$ between the limits for any estimator and the limits for estimators within the hierarchies of convex relaxations. Hence, this result shows statistical optimality must be sacrificed for gaining computational tractability with convex relaxations. For sparse principal submatrix estimation, we prove that a linear-time estimator within $\mathcal{H}$ attains the lower bound in (1.2) up to a logarithmic factor, and is therefore nearly optimal within a general family of convex relaxations.

Our work is closely related to a recent line of research on computational barriers for statistical problems (Berthet and Rigollet, 2013a,b; Ma and Wu, 2013; Krauthgamer et al., 2013; Arias-Castro and Verzelen, 2014; Zhang et al., 2014; Chen and Xu, 2014; Gao et al., 2014; Hajek et al., 2014; Wang et al., 2014; Cai et al., 2015). Under various computational hardness hypotheses on problems like planted clique detection, these works quantify the gap between the information-theoretic limits and the statistical accuracy achievable by polynomial-time algorithms. For this purpose, their proofs are based on polynomial-time reductions from hard computational problems to statistical problems. In contrast with these works, we focus on the statistical limits of a broad class of convex relaxations rather than all polynomial-time algorithms. Correspondingly, our theory does not hinge on unproven computational hardness hypotheses, and our proof is based on constructions rather than reductions. Also, based on another perspective, Chandrasekaran and Jordan (2013) study the tradeoffs between computational complexity and statistical performance for normal mean estimation via hierarchies of convex relaxations. Their results are based on hierarchies of convex constraints, which are obtained by successively weakening the cone representation of the original constraint set. In comparison, our results are based on hierarchies of convex relaxations of the optimization problem itself rather than the constraints, which are obtained by successively tightening a basic semidefinite relaxation using variable augmentation techniques. In addition, our work is connected to previous works on the SoS and other convex relaxation hierarchies (see, e.g., Chlamtac and Tulsiani (2012); Barak and Steurer (2014); Barak and Moitra (2015); Meka et al. (2015) and the references therein). In particular, the key construction of feasible solutions in our proof is based on the dual certificates for the maximum clique problem proposed by Meka et al. (2015).

The rest of this paper is organized as follows. In §2 we introduce the statistical models. In §3 we present the SoS hierarchy of convex relaxations and apply it to estimate the models in §2. In §4 we establish the main results and lay out the proofs in §5. In §6 we conclude the paper.

## 2 Statistical Model

In the sequel, we briefly introduce the statistical models considered in this paper. Then we present several common estimators for them.

### 2.1 Sparse Principal Submatrix Estimation

Let $\mathbf{X} \in \mathbb{R}^{d \times d}$ be a random matrix from distribution $\mathbb{P}$ and $\mathbb{E}(\mathbf{X}) = \mathbf{\Theta}$. We assume there exists an index set $\mathcal{S}^* \subseteq \{1, \ldots, d\}$ with $|\mathcal{S}^*| = s^*$ that satisfies $\Theta_{i,j} = \beta^*$ for $i \neq j$ and $(i,j) \in \mathcal{S}^* \times \mathcal{S}^*$, while $\Theta_{i,j} = 0$ for $i \neq j$ and $(i,j) \notin \mathcal{S}^* \times \mathcal{S}^*$. Here $\beta^* \geq 0$ is the signal strength. For all $i < j$, we assume that $X_{i,j}$'s are independently sub-Gaussian with $\mathbb{E}(X_{i,j}) = \Theta_{i,j}$ and $\|X_{i,j} - \Theta_{i,j}\|_{\psi_2} \leq 1$. In addition, we assume that $X_{i,i} = 0$ and $X_{i,j} = X_{j,i}$. We aim to estimate the signal strength $\beta^*$. For simplicity, hereafter we assume $s^*$ is known. We denote by $\mathcal{P}(s^*, d)$ the family of distribution $\mathbb{P}$'s satisfying the above constraints.

This estimation problem is closely related to the problems considered by Shabalin et al. (2009); Kolar et al. (2011); Butucea and Ingster (2013); Butucea et al. (2013); Ma and Wu (2013); Sun and Nobel (2013); Cai et al. (2015). These works consider the detection problem and the recovery of $\mathcal{S}^*$, while we consider the estimation of signal strength. Besides, we focus on symmetric $\mathbf{X}$ for simplicity.

We consider the following estimator for $\beta^*$ proposed by Butucea and Ingster (2013),

$$\widehat{\beta}^{\text{scan}} = \frac{1}{s^*(s^*-1)} \sup_{\substack{\mathcal{S} \subseteq \{1,\ldots,d\} \\ |\mathcal{S}|=s^*}} \sum_{(i,j) \in \mathcal{S} \times \mathcal{S}} X_{i,j}, \qquad (2.1)$$

where $|\mathcal{S}|$ is the cardinality of set $\mathcal{S}$. The intuition behind $\widehat{\beta}^{\text{scan}}$ is to exhaustively search all principal submatrices of cardinality $s^*$ and calculate the average of all entires within each principal submatrix. In §4 we will prove that $\widehat{\beta}^{\text{scan}}$ attains the information-theoretic lower bound for estimating $\beta^*$ within $\mathcal{P}(s^*, d)$ under a challenging regime where $s^* = o\bigl[(d/\sqrt{\log d})^{2/3}\bigr]$. Nevertheless, it is computationally intractable to obtain $\widehat{\beta}^{\text{scan}}$. In §3 we will introduce convex relaxations of $\widehat{\beta}^{\text{scan}}$. We also consider the following computational tractable estimators

$$\widehat{\beta}^{\text{avg}} = \frac{1}{s^*(s^*-1)} \sum_{i,j=1}^{d} X_{i,j}, \qquad \widehat{\beta}^{\text{max}} = \max_{i,j \in \{1,\ldots,d\}} X_{i,j} \qquad (2.2)$$

for further discussion in §4.

## 2.2 Stochastic Block Model

We consider the estimation of edge probability in a dense subgraph with $s^*$ nodes planted within an Erdős-Rényi graph with $d$ nodes. If a pair of nodes are within the subgraph, they are independently connected with edge probability $\beta^* \in [0, 1]$. Otherwise, they are independently connected with edge probability $\widetilde{\beta}^* \in [0, \beta^*]$. We denote $\mathcal{P}(s^*, d)$ to be the distribution family of graphs which satisfy the above constraints and by $\mathbf{A} \in \mathbb{R}^{d \times d}$ the adjacency matrix. We assume $A_{i,i} = 0$ for all $i \in \{1, \ldots, d\}$ and $s^*$ is known. Similar to principal submatrix estimation, we focus on the challenging regime with $s^* = o\big[(d/\log d)^{2/3}\big]$. Additionally, we assume $\log(d/s^*)/\big(s^*\widetilde{\beta}^*\big) = o(1)$ so that $s^*$ is not too small.

This estimation problem is connected to the problems studied by Kučera (1995); Coja-Oghlan (2010); Bhaskara et al. (2010); Fortunato (2010); Decelle et al. (2011); Mossel et al. (2012, 2013); Verzelen and Arias-Castro (2013); Arias-Castro and Verzelen (2014); Massoulié (2014); Hajek et al. (2014); Chen and Xu (2014); Meka et al. (2015). However, we mainly focus on estimating the edge probability of the dense subgraph rather than detection or recovery of subgraphs. Also, we assume that the dense subgraph and its size are fixed rather than random as in some of the existing works. To estimate $\beta^*$, we use $\widehat{\beta}^{\mathrm{scan}}$ and $\widehat{\beta}^{\max}$ defined in (2.1) and (2.2) with $X_{i,j}$ replaced by $A_{i,j}$. Though stochastic model estimation is closely related to sparse principal submatrix estimation, in §4 we will illustrate that the respective upper and lower bounds have subtle differences because of the different deviations of Bernoulli random variables and general sub-Gaussian random variables, which possibly have unbounded support.

## 3 Convex Relaxation Hierarchy

In this section, we first introduce some specific notations which will greatly simplify our presentation. Then we introduce the SoS hierarchy for $\widehat{\beta}^{\mathrm{scan}}$ defined in (2.1).

**Notation:** We define a collection $\mathcal{C}$ to be an unordered array of elements, where each element can appear more than once. For instance, $\{1\}, \{1, 2\}$ and $\{1, 1\}$ are all collections. Let the summation between two collections be the combination of all elements in them, e.g., for $\mathcal{C}_1 = \{1, 2\}, \mathcal{C}_2 = \{1, 3\}$ we have $\mathcal{C}_1 + \mathcal{C}_2 = \{1, 1, 2, 3\}$. Note that a collection is different from a set, because a set has distinct elements. Let the merge operation $M(\cdot)$ on a collection be the operation that eliminates the duplicate elements and outputs a set, e.g., for $\mathcal{C} = \{1, 1, 2, 2, 3\}$ we have $M(\mathcal{C}) = \{1, 2, 3\}$, which is a set. We use $|\mathcal{C}|$ and $|\mathcal{S}|$ to denote the cardinality of a collection and a set. Also, we denote by $\mathcal{C}_1 = \mathcal{C}_2$ if they contain the same elements. For integer $\ell \geq 0$, we define $d^{(\ell)} = \sum_{i=0}^{\ell} d^i$ for notational simplicity.

Note that $\widehat{\beta}^{\mathrm{scan}}$ in (2.1) can be reformulated as

$$\widehat{\beta}^{\mathrm{scan}} = \max_{\mathbf{v} \in \mathcal{V}_{s^*}} \frac{\mathbf{v}^\top \mathbf{X} \mathbf{v}}{s^*(s^* - 1)}, \quad \text{where } \mathcal{V}_{s^*} = \left\{ \mathbf{v} : \mathbf{v} \in \{0, 1\}^d, \ \sum_{i=1}^{d} v_i = s^* \right\}. \quad (3.1)$$

Because (3.1) involves maximizing a convex function subject to nonconvex constraints, it is computational intractable to solve. Note that $\mathbf{v}^\top \mathbf{X} \mathbf{v} = \mathrm{tr}(\mathbf{X} \mathbf{v} \mathbf{v}^\top)$ in (3.1). We can reparameterize $\mathbf{v} \mathbf{v}^\top$ to be a $d \times d$ positive semidefinite matrix with rank one. For notational simplicity, we define

$$\mathbf{Y} = \begin{bmatrix} 0 & \mathbf{0}_{1\times d} \\ \mathbf{0}_{d\times 1} & \mathbf{X} \end{bmatrix}, \quad \mathbf{\Pi} = (1, \mathbf{v}^\top)^\top (1, \mathbf{v}^\top) = \begin{bmatrix} 1 & \Pi_{0,1} & \cdots & \Pi_{0,d} \\ \Pi_{1,0} & \Pi_{1,1} & \cdots & \Pi_{1,d} \\ \vdots & \vdots & \ddots & \vdots \\ \Pi_{d,0} & \Pi_{d,1} & \cdots & \Pi_{d,d} \end{bmatrix}, \quad v_0 = 1. \quad (3.2)$$

Here $\mathbf{Y}, \mathbf{\Pi} \in \mathbb{R}^{(d+1)\times(d+1)}$ and $\mathbf{0}_{d_1 \times d_2}$ denotes a $d_1 \times d_2$ matrix whose entries are all zero. Meanwhile, note that $\mathcal{V}_{s^*}$ defined in (3.1) can be reformulated as

$$\mathcal{V}_{s^*} = \left\{ \mathbf{v} : \sum_{i=1}^d v_i = s^*, \; v_i^2 - v_i = 0 \text{ for all } i \in \{1, \ldots, d\} \right\}. \quad (3.3)$$

According to the reparametrization in (3.2), it holds that $\Pi_{i,j} = v_i v_j$ for all $i, j \in \{0, \ldots, d\}$. Hence, from (3.1) we obtain the following semidefinite program

$$\max_{\mathbf{\Pi}} \frac{\mathrm{tr}(\mathbf{Y}\mathbf{\Pi})}{s^*(s^*-1)}, \quad \text{subject to } \sum_{i=1}^d \Pi_{i,0} = s^*, \; \Pi_{0,0} = 1, \; \mathbf{\Pi} \succeq \mathbf{0}, \quad (3.4)$$

$$\Pi_{i,j} = \Pi_{j,i}, \; \Pi_{i,i} = \Pi_{i,0} \text{ for all } i, j \in \{0, 1, \ldots, d\},$$

in which $\sum_{i=1}^d \Pi_{i,0} = s^*$ corresponds to $\sum_{i=1}^d v_i = s^*$, $\Pi_{i,j} = \Pi_{j,i}$ corresponds to $v_i v_j = v_j v_i$, while $\Pi_{i,i} = \Pi_{i,0}$ corresponds to $v_i^2 - v_i = 0$. Note that if $\mathrm{rank}(\mathbf{\Pi}) = 1$, then from our reparametrization in (3.2), the maximum of (3.4) equals the maximum of (3.1). However, we drop this rank constraint since it is nonconvex, and hence (3.4) is a convex relaxation of (3.1).

The SoS hierarchy is obtained by increasingly tightening the basic semidefinite program in (3.4) using variable augmentation techniques. In particular, the reparametrization in (3.4) only involves the second order interaction between $v_i$ and $v_j$. For integer $\ell \geq 1$, we consider a $d^{(\ell)} \times d^{(\ell)}$ matrix $\mathbf{\Pi}^{(\ell)}$, where $d^{(\ell)} = \sum_{i=0}^\ell d^i$ in our notations. For notational simplicity, we index the entries of $\mathbf{\Pi}^{(\ell)}$ using collections $\mathcal{C}_1$ and $\mathcal{C}_2$ with $|\mathcal{C}_1|, |\mathcal{C}_2| \leq \ell$, whose elements are indices $1, \ldots, d$. Our reparametrization takes the form

$$\Pi^{(\ell)}_{\mathcal{C}_1, \mathcal{C}_2} = \prod_{i \in \mathcal{C}_1} v_i \prod_{j \in \mathcal{C}_2} v_j = \prod_{i \in \mathcal{C}_1 + \mathcal{C}_2} v_i. \quad (3.5)$$

In particular, for $\mathcal{C} = \varnothing$ we define $\prod_{i \in \mathcal{C}} v_i = 1$. The $\ell$-th level SoS relaxation of (3.1) takes the form

$$\max_{\mathbf{\Pi}} \frac{\mathrm{tr}\big(\mathbf{Y}^{(\ell)} \mathbf{\Pi}^{(\ell)}\big)}{s^*(s^*-1)}, \quad \text{subject to } \sum_{i=1}^d \Pi^{(\ell)}_{\{i\}+\mathcal{C}_1, \mathcal{C}_2} = s^* \Pi^{(\ell)}_{\mathcal{C}_1, \mathcal{C}_2}, \text{ for all } |\mathcal{C}_1| \leq \ell-1, \; |\mathcal{C}_2| \leq \ell, \quad (3.6)$$

$$\Pi^{(\ell)}_{\{i,i\}+\mathcal{C}_1, \mathcal{C}_2} = \Pi^{(\ell)}_{\{i\}+\mathcal{C}_1, \mathcal{C}_2}, \text{ for all } i \in \{1, \ldots, d\}, \; |\mathcal{C}_1| \leq \ell-2, \; |\mathcal{C}_2| \leq \ell,$$

$$\Pi^{(\ell)}_{\mathcal{C}_1, \mathcal{C}_2} = \Pi^{(\ell)}_{\mathcal{C}'_1, \mathcal{C}'_2}, \text{ for all } \mathcal{C}_1 + \mathcal{C}_2 = \mathcal{C}'_1 + \mathcal{C}'_2, \; |\mathcal{C}_1|, |\mathcal{C}_2|, |\mathcal{C}'_1|, |\mathcal{C}'_2| \leq \ell,$$

$$\Pi^{(\ell)}_{\varnothing, \varnothing} = 1, \; \mathbf{\Pi}^{(\ell)} \succeq \mathbf{0},$$

where $\mathbf{Y}^{(\ell)} \in \mathbb{R}^{d^{(\ell)} \times d^{(\ell)}}$ is defined as

$$\mathbf{Y}^{(\ell)} = \begin{bmatrix} 0 & \mathbf{0}_{1 \times d} & \cdots & \mathbf{0}_{1 \times d^\ell} \\ \mathbf{0}_{d \times 1} & \mathbf{X} & \cdots & \mathbf{0}_{d \times d^\ell} \\ \vdots & \vdots & \ddots & \vdots \\ \mathbf{0}_{d^\ell \times 1} & \mathbf{0}_{d^\ell \times d} & \cdots & \mathbf{0}_{d^\ell \times d^\ell} \end{bmatrix}.$$

In (3.6), the first constraint corresponds to the reparametrization in (3.5) and

$$\prod_{j \in \mathcal{C}} v_j \left( \sum_{i=1}^d v_i \right) = s^* \prod_{j \in \mathcal{C}} v_j, \quad \text{for all } |\mathcal{C}| \leq 2\ell - 1,$$

which is equivalent to $\sum_{i=1}^d v_i = s^*$ in (3.3). The second constraint corresponds to (3.5) and

$$\prod_{j \in \mathcal{C}} v_j \cdot v_i^2 = \prod_{j \in \mathcal{C}} v_j \cdot v_i, \quad \text{for all } |\mathcal{C}| \leq 2\ell - 2,$$

which is equivalent to $v_i^2 - v_i = 0$ in (3.3). The third constraint corresponds to (3.5) and

$$\prod_{j \in \mathcal{C}_1 + \mathcal{C}_2} v_j = \prod_{j \in \mathcal{C}_1' + \mathcal{C}_2'} v_j, \quad \text{for all } \mathcal{C}_1 + \mathcal{C}_2 = \mathcal{C}_1' + \mathcal{C}_2', \ |\mathcal{C}_1|, |\mathcal{C}_2|, |\mathcal{C}_1'|, |\mathcal{C}_2'| \leq \ell.$$

The last constraint that $\Pi_{\varnothing,\varnothing}^{(\ell)} = 1$ follows from (3.5) and our definition that $\prod_{i \in \mathcal{C}} v_i = 1$ for $\mathcal{C} = \varnothing$. For $\ell = 1$, (3.6) reduces to the basic semidefinite relaxation in (3.4). We denote by $\widehat{\beta}_{\text{SoS}}^{(\ell)}$ the maximum of (3.6). We have

$$\widehat{\beta}^{\text{scan}} \leq \cdots \leq \widehat{\beta}_{\text{SoS}}^{(\ell)} \leq \cdots \leq \widehat{\beta}_{\text{SoS}}^{(2)} \leq \widehat{\beta}_{\text{SoS}}^{(1)},$$

since we have more constraints in (3.6) for a larger $\ell$. Thus, for a larger $\ell$ (3.6) gives a tighter convex relaxation of (3.1). Meanwhile, note that the semidefinite program in (3.6) can be solved in $O(d^{O(\ell)})$ operations. Hereafter we focus on the settings where $\ell$ does not increase with $d$.

Laurent (2003) proves that other existing convex relaxation hierarchies, such as Sherali-Adams and Lovász-Schrijver hierarchies as well as their extensions, are weaker than the SoS hierarchy in the sense that $\widehat{\beta}^{\text{scan}} \leq \widehat{\beta}_{\text{SoS}}^{(\ell)} \leq \widehat{\beta}_{\text{other}}^{(\ell)}$, where $\widehat{\beta}_{\text{other}}^{(\ell)}$ denotes the $\ell$-th level of other weaker hierarchies. Note that relaxing constraints and objectives in the convex relaxations also leads to looser approximations of $\widehat{\beta}^{\text{scan}}$. Hence, we denote by $\mathcal{H}^{(\ell)}$ the class of estimator $\widehat{\beta}$'s that fall in the $\ell$-th level of the SoS and weaker hierarchies, as well as their weakened versions obtained by relaxing constraints and objectives. By this definition, we have $\mathcal{H}^{(1)} \subseteq \mathcal{H}^{(2)} \cdots$. For example, for $\ell > 1$ we can drop constraints in (3.6) to obtain (3.4), which corresponds to $\ell = 1$. In particular, from (3.1) we have

$$\widehat{\beta}^{\text{scan}} = \max_{\mathbf{v} \in \mathcal{V}_{s^*}} \frac{\mathbf{v}^\top \mathbf{X} \mathbf{v}}{s^*(s^* - 1)} \leq \max_{\mathbf{u}, \mathbf{v} \in \mathcal{V}_{s^*}} \frac{\mathbf{u}^\top \mathbf{X} \mathbf{v}}{s^*(s^* - 1)} \leq \max_{\mathbf{u}, \mathbf{v} \in \overline{\mathcal{V}}_{s^*}} \frac{\mathbf{u}^\top \mathbf{X} \mathbf{v}}{s^*(s^* - 1)} \leq \max_{\mathbf{\Omega} \in \mathcal{W}_{s^*}} \frac{\text{tr}(\mathbf{X} \mathbf{\Omega})}{s^*(s^* - 1)}, \quad (3.7)$$

$$\text{where } \overline{\mathcal{V}}_{s^*} = \left\{ \mathbf{v} : \sum_{i=1}^d v_i = s^*, \ v_i \geq 0 \text{ for all } i \in \{1, \ldots, d\} \right\},$$

$$\mathcal{W}_{s^*} = \left\{ \mathbf{\Omega} : \sum_{i=1}^d \Omega_{i,j} = (s^*)^2, \ \Omega_{i,j} \geq 0 \text{ for all } i, j \in \{1, \ldots, d\} \right\}.$$

Here $\overline{\mathcal{V}}_{s^*}$ is a linear relaxation of $\mathcal{V}_{s^*}$. Note that the right-hand side of (3.7) equals $s^*/(s^*-1) \cdot \widehat{\beta}^{\mathrm{max}}$, where $\widehat{\beta}^{\mathrm{max}}$ is defined in (2.2). Therefore, $s^*/(s^*-1) \cdot \widehat{\beta}^{\mathrm{max}}$ can be viewed as a linear programming relaxation of $\widehat{\beta}^{\mathrm{scan}}$, which falls within $\mathcal{H}^{(1)}$ (see, e.g., §2 of Chlamtac and Tulsiani (2012) for details). In addition, it is worth noting that the SoS hierarchy has several equivalent formulations. See, e.g., Theorem 2.7 of Barak and Steurer (2014) for a proof of such equivalence.

## 4 Main Result

As defined in §3, $\mathcal{H}^{(\ell)}$ denotes the $\ell$-th level of the convex relaxation hierarchy for $\widehat{\beta}^{\mathrm{scan}}$ defined in (2.1). For stochastic block model, we replace $\mathbf{X}$ in (2.1) with the adjacency matrix $\mathbf{A}$ respectively.

### 4.1 Sparse Principal Submatrix Estimation

In the following, we present the main theoretical results for estimating the signal strength of sparse principal submatrix. In the sequel we establish the information-theoretic lower bound for estimating $\beta^*$ within the distribution family $\mathcal{P}(s^*, d)$ defined in §2.1.

**Theorem 4.1.** For all estimators $\widehat{\beta}$ constructed using $\mathbf{X} \sim \mathbb{P} \in \mathcal{P}(s^*, d)$ and $s^* = o\big[(d/\sqrt{\log d})^{2/3}\big]$, there exists an absolute constant $C > 0$ such that

$$\inf_{\widehat{\beta}} \sup_{\mathbb{P} \in \mathcal{P}(s^*,d)} \mathbb{E}_{\mathbb{P}} \big|\widehat{\beta} - \beta^*\big| \geq C \sqrt{1/s^* \cdot \log(d/s^*)}.$$

*Proof.* See §5.1 for a detailed proof. □

In Theorem 4.1 we consider a challenging regime. More specifically, a straightforward calculation shows that $\widehat{\beta}^{\mathrm{avg}}$ defined in (2.2) achieves the $d/(s^*)^2$ rate of convergence. For $s^* = o\big[(d/\sqrt{\log d})^{2/3}\big]$, we have $\sqrt{1/s^* \cdot \log(d/s^*)} = o\big[d/(s^*)^2\big]$. Thus, there exists a gap between the rate attained by $\widehat{\beta}^{\mathrm{avg}}$ and the information-theoretic lower bound. We will show that there is also such a gap for $\widehat{\beta}^{\mathrm{max}}$. The next proposition shows $\widehat{\beta}^{\mathrm{scan}}$ in (2.1) attains the information-theoretic lower bound in Theorem 4.1.

**Proposition 4.2.** For $\widehat{\beta}^{\mathrm{scan}}$ defined in (2.1) with $X_{i,j}$ being the $(i,j)$-th entry of $\mathbf{X} \sim \mathbb{P} \in \mathcal{P}(s^*, d)$, we have that

$$\big|\widehat{\beta}^{\mathrm{scan}} - \beta^*\big| \leq C\sqrt{1/s^* \cdot \log(d/s^*)}$$

holds with probability at least $1 - 1/d$ for some absolute constant $C > 0$.

*Proof.* See §5.1 for a detailed proof. □

Theorem 4.1 and Proposition 4.2 show that $\widehat{\beta}^{\mathrm{scan}}$ is statistically optimal under the regime where $s^* = o\big[(d/\sqrt{\log d})^{2/3}\big]$. However, it is computationally intractable to obtain $\widehat{\beta}^{\mathrm{scan}}$. Thus, we consider the family of convex relaxations of $\widehat{\beta}^{\mathrm{scan}}$ within the $\ell$-th level SoS and weaker hierarchies as well as their further relaxations, which is denoted by $\mathcal{H}^{(\ell)}$. In the sequel, we establish a minimax lower bound for the statistical performance of all estimators within $\mathcal{H}^{(\ell)}$. Recall that $\mathcal{P}(s^*, d)$ is the distribution family defined in §2.1.

**Theorem 4.3.** We assume $s^* = o\{[d/(\log d)^2]^{1/2\ell}\}$. There is an absolute constant $C > 0$ such that

$$\inf_{\widehat{\beta} \in \mathcal{H}^{(\ell)}} \sup_{\mathbb{P} \in \mathcal{P}(s^*, d)} \mathbb{E}_{\mathbb{P}} |\widehat{\beta} - \beta^*| \geq C.$$

*Proof.* See §5.1 for a detailed proof. □

Note that the regime considered in Theorem 4.3 is within the challenging regime considered in Theorem 4.1. Under this regime, Theorem 4.3 proves that any estimator within the convex relaxation hierarchy fails to attain a statistical rate that decreases when $s^*$ is increasing. A comparison between Theorems 4.1 and 4.3 illustrates that there exists a gap of $\sqrt{s^*}$ (ignoring the $\log d$ factor) between the information-theoretic lower bound and the statistical rate achievable by a broad class of convex relaxations. In other words, to achieve computational tractability via convex relaxations, we have to compromise statistical optimality.

It is worth noting that this gap between computational tractability and statistical optimality is effective under the regime $s^* = o\{[d/(\log d)^2]^{1/2\ell}\}$, which shrinks as $\ell$ increases. However, $\ell$ cannot increase with $d$ and $s^*$, because otherwise the computational complexity required to solve the convex relaxations increases exponentially, according to our discussion in §3. For $\ell$ being any constant, the regime in Theorem 4.3 is a nontrivial subset of the regime in Theorem 4.1. As will be shown in our proof, $s^* = o\{[d/(\log d)^2]^{1/2\ell}\}$ is a sufficient condition to establish the feasibility of the constructed solution. In fact, for $\ell = 2$, we can further relax this condition to $s^* = o(d^{1/3}/\log d)$ with the results of Deshpande and Montanari (2015). Under the regime in Theorem 4.3, the next proposition shows that $\widehat{\beta}^{\max}$ defined in (2.2) is nearly optimal under computational tractability constraints.

**Proposition 4.4.** For $\widehat{\beta}^{\max}$ in (2.2), where $X_{i,j}$ is the $(i,j)$-th entry of $\mathbf{X} \sim \mathbb{P} \in \mathcal{P}(s^*, d)$, we have

$$|\widehat{\beta}^{\max} - \beta^*| \leq C\sqrt{\log d}$$

holds with probability at least $1 - 1/d$ for some absolute constant $C > 0$.

*Proof.* See §5.1 for a detailed proof. □

According to (3.7) and the discussion in §3, we have $\widehat{\beta}^{\max} \in \mathcal{H}^{(1)} \subseteq \mathcal{H}^{(2)} \cdots$. Thus $\widehat{\beta}^{\max}$ attains the minimax lower bound with computational constraints in Theorem 4.3 for every $\ell$ up to a $\log d$ factor, which also suggests that the lower bound in Theorem 4.3 is tight. Meanwhile, note that the calculation of $\widehat{\beta}^{\max}$ in (2.2) requires $O(d^2)$ operations, which is linear in the size of input. In contrast, tighter approximations in the $\ell$-th level SoS hierarchy require $O(d^{O(\ell)})$ operations. In practice, such a computational complexity is in general higher than the complexity for calculating $\widehat{\beta}^{\max}$. Theorem 4.3 indicates that this extra computational cost can only result in limited possible improvements on the statistical rate of convergence, i.e., a $\log d$ factor.

It is worth noting the gap between the lower bounds in Theorems 4.1 and 4.3 vanishes when $s^*$ is a constant that does not increase with $d$. In this case, $\widehat{\beta}^{\max}$ achieves the information-theoretic lower bound in Theorem 4.1. On the other hand, $\widehat{\beta}^{\text{scan}}$ is computational tractable to obtain in this case.

## 4.2 Stochastic Block Model

In this section, we present the main theory for edge probability estimation in stochastic block model. Recall that $\mathcal{P}(s^*, d)$ is the distribution family defined in §2.2. The following lemma establishes the information-theoretic lower bound for estimating $\beta^*$. Recall $\widetilde{\beta}^*$ denotes the edge probability of the large Erdős-Rényi graph with $d$ nodes.

**Theorem 4.5.** For $s^* = o\big[(d/\sqrt{\log d})^{2/3}\big]$ and $\log(d/s^*)/(s^*\widetilde{\beta}^*) = o(1)$, there is an absolute constant $C > 0$ such that
$$\inf_{\widehat{\beta}} \sup_{\mathbb{P} \in \mathcal{P}(s^*, d)} \mathbb{E}_{\mathbb{P}}\big|\widehat{\beta} - \beta^*\big| \geq C\sqrt{1/s^* \cdot \log(d/s^*)}.$$

*Proof.* See §5.2 for a detailed proof. □

Theorem 4.5 is similar to Theorem 4.1 but needs an extra condition that $\log(d/s^*)/(s^*\widetilde{\beta}^*) = o(1)$, which ensures $s^*$ is not too small. Recall each entry of the adjacency matrix **A** is Bernoulli. Arias-Castro and Verzelen (2014) shows that a larger $s^*$ guarantees the moderate deviation of the Bernoulli distribution is in effect in the lower bound. Next, we prove $\widehat{\beta}^{\text{scan}}$ achieves the information-theoretic lower bound in Theorem 4.5 and hence is optimal.

**Proposition 4.6.** For $\widehat{\beta}^{\text{scan}}$ defined in (2.1), we have that with probability at least $1 - 1/d$,
$$\big|\widehat{\beta}^{\text{scan}} - \beta^*\big| \leq C\sqrt{1/s^* \cdot \log(d/s^*)}.$$

*Proof.* See §5.2 for a detailed proof. □

The next theorem establishes the minimax lower bound on the statistical performance of convex relaxations within $\mathcal{H}^{(\ell)}$ defined in §3.

**Theorem 4.7.** For $s^*$ and $d$ sufficiently large and $s^* = o\big\{[d/(\log d)^2]^{1/2\ell}\big\}$, we have
$$\inf_{\widehat{\beta} \in \mathcal{H}^{(\ell)}} \sup_{\mathbb{P} \in \mathcal{P}(s^*, d)} \mathbb{E}_{\mathbb{P}}\big|\widehat{\beta} - \beta^*\big| \geq 1/4.$$

*Proof.* See §5.2 for a detailed proof. □

Similar to Theorem 4.3, Theorem 4.7 shows the gap between statistical optimality and computational tractability. Note that $\beta^* \in [0, 1]$. Meanwhile, it is easy to show $\mathbb{P}\big(\widehat{\beta}^{\max} = 1\big) \geq 1 - \big(1 - \widetilde{\beta}^*\big)^{\frac{d^2-d}{2}}$. Therefore, $\widehat{\beta}^{\max}$ exactly attains such a minimax lower bound under computational constraints up to constants. From another point of view, for $s^* = o\big\{[d/(\log d)^2]^{1/2\ell}\big\}$, every estimators within $\mathcal{H}^{(\ell)}$ is at most as accurate as the trivial estimator $\widehat{\beta} = 1$.

Theorems 4.3 and 4.7 are similar. Note that for sparse principal submatrix estimation we consider sub-Gaussian entries, while in the adjacency matrix for stochastic block model each entry is Bernoulli. A direct way to establish Theorem 4.3 is to adapt the construction of $\mathbb{P}$ in the proof of Theorem 4.7, since Bernoulli is sub-Gaussian. However, as illustrated in §5.1 the information-theoretic lower bound in Theorem 4.1 is established using the construction of $\mathbb{P}$ with unbounded support. Correspondingly, we use a construction of $\mathbb{P}$ with unbounded support to establish the lower bound with computational constraints in Theorem 4.3. By matching the constructions of $\mathbb{P} \in \mathcal{P}(s^*, d)$ in the proofs of Theorems 4.1 and 4.3, we can sharply characterize the existence of the $\sqrt{s^*}$ gap particularly for sub-Gaussian distributions with unbounded support.

## 5 Proof of Main Results

In the sequel, we present the proofs of the main results in §4. We first lay out the proofs for sparse principal submatrix estimation, and then the proofs for stochastic block model.

### 5.1 Proof for Sparse Principal Submatrix Estimation

Before we establish the proof of Theorem 4.1, we present a corollary of Theorem 2.2 of Butucea and Ingster (2013). Let $\bar{\beta}$ be a quantity that scales with $s^*$ and $d$. It establishes the sufficient conditions under which distinguishing $\beta^* = 0$ and $\beta^* = \bar{\beta}$ is impossible. Recall $\mathcal{P}(s^*, d)$ denotes the distribution family specified in §2.1.

**Corollary 5.1.** We consider testing $H_0 : \beta_0^* = 0$ against $H_1 : \beta_1^* = \bar{\beta}$. For any test $\phi : \mathbb{R}^{d \times d} \to \{0, 1\}$ based on $\mathbf{X}$, if $\bar{\beta}^2 (s^*)^4 / d^2 = o(1)$ and $\limsup \bar{\beta}^2 s^* / \log(d/s^*) < C$, there exist $\mathbb{P}_0, \mathbb{P}_1 \in \mathcal{P}(s^*, d)$, which correspond to $H_0$ and $H_1$, such that

$$\inf_{\phi} \max\{\mathbb{P}_0(\phi = 1),\ \mathbb{P}_1(\phi = 0)\} \geq 1/4.$$

Here $C > 0$ is an absolute constant.

*Proof.* Theorem 2.2 of Butucea and Ingster (2013) gives a similar result for $\mathbf{X}$ with Gaussian entries. Therefore, their $\mathbb{P}_0$ and $\mathbb{P}_1$ fall within $\mathcal{P}(s^*, d)$ specified in §2.1 up to rescaling of variance. Besides, it is worth noting that Butucea and Ingster (2013) do not assume $\mathbf{X}$ is symmetric. Nevertheless, the proof for symmetric $\mathbf{X}$ follows similarly from their proof. □

Equipped with Corollary 5.1, we are now ready to prove Theorem 4.1.

*Proof of Theorem 4.1.* We consider testing hypotheses $H_0 : \beta_0^* = 0$ and $H_1 : \beta_1^* = \bar{\beta}$ with

$$\bar{\beta} = C\sqrt{1/s^* \cdot \log(d/s^*)}, \tag{5.1}$$

where $C$ is an absolute constant that is sufficiently small. By Corollary 5.1, there exist $\mathbb{P}_0, \mathbb{P}_1 \in \mathcal{P}(s^*, d)$ corresponding to $H_0$ and $H_1$, such that for any test $\phi : \mathbb{R}^{d \times d} \to \{0, 1\}$,

$$\inf_{\phi} \max\{\mathbb{P}_0(\phi = 1),\ \mathbb{P}_1(\phi = 0)\} \geq 1/4, \quad \text{for } \bar{\beta}(s^*)^2/d = o(1). \tag{5.2}$$

We consider a specific test $\bar\phi(\widehat\beta)$ based on $\widehat\beta$, which is defined as $\bar\phi(\widehat\beta) = \mathbb{1}(\widehat\beta > \bar\beta/2)$. From (5.2) we have

$$\inf_{\widehat\beta} \max\Big\{\mathbb{P}_0\big(|\widehat\beta - \beta_0^*| \geq \bar\beta/2\big),\ \mathbb{P}_1\big(|\widehat\beta - \beta_1^*| \geq \bar\beta/2\big)\Big\}$$

$$= \inf_{\widehat\beta} \max\Big\{\mathbb{P}_0\big(|\widehat\beta| \geq \bar\beta/2\big),\ \mathbb{P}_1\big(|\widehat\beta - \bar\beta| \geq \bar\beta/2\big)\Big\}$$

$$\geq \inf_{\widehat\beta} \max\Big\{\mathbb{P}_0\big[\bar\phi(\widehat\beta) = 1\big],\ \mathbb{P}_1\big[\bar\phi(\widehat\beta) = 0\big]\Big\} \geq \inf_{\phi} \max\big\{\mathbb{P}_0(\phi = 1),\ \mathbb{P}_1(\phi = 0)\big\} \geq 1/4. \quad (5.3)$$

Here the first inequality holds because under $H_0$, $\bar\phi(\widehat\beta) = 1$ implies $|\widehat\beta - \beta_0^*| \geq \bar\beta/2$ by definition and under $H_1$, $\bar\phi(\widehat\beta) = 0$ implies $|\widehat\beta - \beta_1^*| \geq \bar\beta/2$. Here the second last inequality holds because $\bar\phi(\widehat\beta)$ is a specific class of tests. Consequently, we have

$$\inf_{\widehat\beta} \sup_{\mathbb{P} \in \mathcal{P}(s^*, d)} \mathbb{E}_{\mathbb{P}} |\widehat\beta - \beta^*| \geq \inf_{\widehat\beta} \max\Big\{\mathbb{E}_{\mathbb{P}_0}|\widehat\beta - \beta_0^*|,\ \mathbb{E}_{\mathbb{P}_1}|\widehat\beta - \beta_1^*|\Big\}$$

$$\geq \bar\beta/2 \cdot \inf_{\widehat\beta} \max\Big\{\mathbb{P}_0\big(|\widehat\beta - \beta_0^*| \geq \bar\beta/2\big),\ \mathbb{P}_1\big(|\widehat\beta - \beta_1^*| \geq \bar\beta/2\big)\Big\} \geq \bar\beta/8, \quad (5.4)$$

where the second inequality is from Markov's inequality and the last is from (5.3). By plugging (5.1) into (5.4), we reach the conclusion. $\square$

In the sequel, we prove the upper bound in Proposition 4.2.

*Proof of Proposition 4.2.* For integer $s > 0$, we denote by $\mathcal{V}_s$ the set of $\mathbf{v} \in \mathbb{R}^d$ with exactly $s$ entries being one and the others being zero. By definition, in (2.1) we have

$$\sup_{\substack{\mathcal{S} \subseteq \{1,\ldots,d\} \\ |\mathcal{S}| = s^*}} \sum_{(i,j) \in \mathcal{S} \times \mathcal{S}} X_{i,j} = \sup_{\mathbf{v} \in \mathcal{V}_{s^*}} \mathbf{v}^\top \mathbf{X} \mathbf{v}/2. \quad (5.5)$$

Recall that by our definition we have $X_{i,i} = 0$ for all $i \in \{1, \ldots, d\}$ and $\mathbb{E}\mathbf{X} = \boldsymbol{\Theta}$. Note that

$$\Big| \sup_{\mathbf{v} \in \mathcal{V}_{s^*}} \mathbf{v}^\top \mathbf{X} \mathbf{v} - \sup_{\mathbf{v} \in \mathcal{V}_{s^*}} \mathbf{v}^\top \boldsymbol{\Theta} \mathbf{v} \Big| \leq \sup_{\mathbf{v} \in \mathcal{V}_{s^*}} \big|\mathbf{v}^\top (\mathbf{X} - \boldsymbol{\Theta}) \mathbf{v}\big|. \quad (5.6)$$

Since $\mathbf{X} \sim \mathbb{P} \in \mathcal{P}(s^*, d)$, for any fixed $\mathbf{v} \in \mathcal{V}_{s^*}$, $\mathbf{v}^\top (\mathbf{X} - \boldsymbol{\Theta}) \mathbf{v}$ is twice the summation of $s^*(s^* - 1)/2$ independent sub-Gaussian random variables that have mean zero and $\psi_2$-norm at most one. Hence, for any fixed $\mathbf{v} \in \mathcal{V}_{s^*}$ we have

$$\mathbb{P}\big[\big|\mathbf{v}^\top (\mathbf{X} - \boldsymbol{\Theta}) \mathbf{v}\big| > t\big] < \exp\big\{1 - Ct^2/[s^*(s^* - 1)]\big\}.$$

Then by union bound, we have

$$\mathbb{P}\Big[\sup_{\mathbf{v} \in \mathcal{V}_{s^*}} \big|\mathbf{v}^\top (\mathbf{X} - \boldsymbol{\Theta}) \mathbf{v}\big| > t\Big] \leq \binom{d}{s^*} \exp\big\{1 - Ct^2/[s^*(s^* - 1)]\big\}$$

$$\leq \exp\big\{1 - Ct^2/[s^*(s^* - 1)] + s^* \log(d/s^*)\big\}.$$

Setting the right-hand side to be $\delta$, we obtain

$$t = C\sqrt{\log(e/\delta) + s^* \log(d/s^*)} \cdot \sqrt{s^*(s^* - 1)}. \tag{5.7}$$

Plugging (5.7) into (5.6), we have that with probability at least $1 - \delta$,

$$\left| \sup_{\mathbf{v} \in \mathcal{V}_{s^*}} \mathbf{v}^\top \mathbf{X} \mathbf{v} - \sup_{\mathbf{v} \in \mathcal{V}_{s^*}} \mathbf{v}^\top \mathbf{\Theta} \mathbf{v} \right| \leq C\sqrt{\log(e/\delta) + s^* \log(d/s^*)} \cdot \sqrt{s^*(s^* - 1)}.$$

Note that $\sup_{\mathbf{v} \in \mathcal{V}_{s^*}} \mathbf{v}^\top \mathbf{\Theta} \mathbf{v} = s^*(s^* - 1) \cdot \beta^*$. Then by (2.1) and (5.5) we obtain that

$$\left| \widehat{\beta}^{\text{scan}} - \beta^* \right| \leq C\sqrt{\log(e/\delta) + s^* \log(d/s^*)} / \sqrt{s^*(s^* - 1)}$$

holds with probability at least $1 - \delta$. Setting $\delta = 1/d$, we reach the conclusion. $\square$

In the following we prove Theorem 4.3.

*Proof of Theorem 4.3.* In this proof, we focus on specific distributions in $\mathcal{P}(s^*, d)$ with $\beta^* = 0$. We consider $X_{i,j}$'s ($i < j$) being sub-Gaussian random variables which satisfy the constraints in §2.1. In addition, we assume that $|X_{i,j}| \geq \nu$ almost surely and $\mathbb{P}(X_{i,j} > 0) = \mathbb{P}(X_{i,j} < 0) = 1/2$ for all $i < j$ and constant $\nu > 0$. Under such a distribution, we construct a matrix $\mathbf{\Pi}^{(\ell)} \in \mathbb{R}^{d^{(\ell)} \times d^{(\ell)}}$, which is a feasible solution to the $\ell$-th level SoS program in (3.6) with high probability. We further prove that the objective value corresponding to $\mathbf{\Pi}^{(\ell)}$ is larger than $\nu$, which indicates that the maximum of the corresponding SoS program is at least $\nu$ with high probability. In the rest of this proof, we denote $\mathbf{X} + \nu \cdot \mathbf{I}_d$ to be $\overline{\mathbf{X}}$.

Hereafter, we denote by $\overline{\mathbf{X}}_{\mathcal{S}, \mathcal{S}'}$ the submatrix of $\overline{\mathbf{X}}$ whose row indices are in $\mathcal{S}$ and column indices are in $\mathcal{S}'$. For notational simplicity, we define the expansivity $\eta(\mathcal{S}, \overline{\mathbf{X}})$ of some set $\mathcal{S} \subseteq \{1, \ldots, d\}$ to be the number of sets $\mathcal{S}' \subseteq \{1, \ldots, d\}$ that satisfy $|\mathcal{S}'| = 2\ell$, $\mathcal{S} \subseteq \mathcal{S}'$ and $\text{sign}(\overline{\mathbf{X}}_{\mathcal{S}', \mathcal{S}'}) = \mathbf{1}_{2\ell, 2\ell}$. Here $\text{sign}(\mathbf{X})$ is a matrix that satisfies $[\text{sign}(\mathbf{X})]_{i,j} = 1$ if $X_{i,j} > 0$ and $[\text{sign}(\mathbf{X})]_{i,j} = 0$ if $X_{i,j} \leq 0$. Note that $\eta(\mathcal{S}, \overline{\mathbf{X}})$ is nonzero only if $\overline{X}_{i,j} > 0$ for all $i \in \mathcal{S}, j \in \mathcal{S}$. Hence, $\eta(\mathcal{S}, \overline{\mathbf{X}})$ gives the number of $\overline{\mathbf{X}}$'s submatrices that are extended from $\overline{\mathbf{X}}_{\mathcal{S}, \mathcal{S}}$ and have size $2\ell \times 2\ell$ with all entries being positive. It is worth noting that by definition $\eta(\mathcal{S}, \overline{\mathbf{X}})$ is a random quantity, which depends on the random matrix $\overline{\mathbf{X}}$. Recall that each entry $\Pi^{(\ell)}_{\mathcal{C}_1, \mathcal{C}_2}$ of $\mathbf{\Pi}^{(\ell)}$ are indexed by collections $\mathcal{C}_1$ and $\mathcal{C}_2$, and $M(\mathcal{C}_1)$ and $M(\mathcal{C}_1)$ are the respective sets, which have distinct elements. Based on the construction of dual certificates of Meka et al. (2015), we construct $\mathbf{\Pi}^{(\ell)}$ as

$$\Pi^{(\ell)}_{\mathcal{C}_1, \mathcal{C}_2} = \frac{\eta\big[M(\mathcal{C}_1) \cup M(\mathcal{C}_2), \overline{\mathbf{X}}\big]}{\eta(\varnothing, \overline{\mathbf{X}})} \cdot \frac{s^*!/[s^* - |M(\mathcal{C}_1) \cup M(\mathcal{C}_2)|]!}{(2\ell)!/[2\ell - |M(\mathcal{C}_1) \cup M(\mathcal{C}_2)|]!}. \tag{5.8}$$

Now we verify $\mathbf{\Pi}^{(\ell)}$ defined in (5.8) satisfies all the constraints of the $\ell$-th level SoS program in (3.6). First, we have $\Pi^{(\ell)}_{\varnothing, \varnothing} = 1$ from (5.8). Also, $\mathbf{\Pi}^{(\ell)}$ satisfies $\Pi^{(\ell)}_{\mathcal{C}_1, \mathcal{C}_2} = \Pi^{(\ell)}_{\mathcal{C}'_1, \mathcal{C}'_2}$ for $\mathcal{C}_1 + \mathcal{C}_2 = \mathcal{C}'_1 + \mathcal{C}'_2$, since

$$M(\mathcal{C}_1) \cup M(\mathcal{C}_2) = M(\mathcal{C}_1 + \mathcal{C}_2) = M(\mathcal{C}'_1 + \mathcal{C}'_2) = M(\mathcal{C}'_1) \cup M(\mathcal{C}'_2)$$

by the definition of the merge operation $M(\cdot)$. Meanwhile, it holds that $\Pi^{(\ell)}_{\mathcal{C}_1+\{i,i\},\mathcal{C}_2} = \Pi^{(\ell)}_{\mathcal{C}_1+\{i\},\mathcal{C}_2}$ for all $\mathcal{C}_1$ and $\mathcal{C}_2$ with $|\mathcal{C}_1| \leq \ell - 2$ and $|\mathcal{C}_2| \leq \ell$, since in (5.8) we have

$$M(\mathcal{C}_1 + \{i,i\}) \cup M(\mathcal{C}_2) = M(\mathcal{C}_1 + \{i\}) \cup M(\mathcal{C}_2).$$

Now we prove that $\sum_{i=1}^d \Pi^{(\ell)}_{\mathcal{C}_1+\{i\},\mathcal{C}_2} = s^* \Pi^{(\ell)}_{\mathcal{C}_1,\mathcal{C}_2}$ holds for all $|\mathcal{C}_1| \leq \ell - 1$ and $|\mathcal{C}_2| \leq \ell$. Let $\mathcal{C} = \mathcal{C}_1 + \mathcal{C}_2$, which satisfies $|M(\mathcal{C})| \leq |\mathcal{C}| \leq 2\ell - 1$. By (5.8) we have

$$\sum_{i=1}^d \Pi^{(\ell)}_{\mathcal{C}_1+\{i\},\mathcal{C}_2} = \sum_{i=1}^d \frac{\eta\big[M(\mathcal{C}+\{i\}),\overline{\mathbf{X}}\big]}{\eta(\varnothing,\overline{\mathbf{X}})} \cdot \frac{s^*!/[s^* - |M(\mathcal{C}+\{i\})|]!}{(2\ell)!/[2\ell - |M(\mathcal{C}+\{i\})|]!}, \tag{5.9}$$

where we use the fact that

$$M(\mathcal{C}_1 + \{i\}) \cup M(\mathcal{C}_2) = M(\mathcal{C}_1 + \mathcal{C}_2 + \{i\}) = M(\mathcal{C} + \{i\}).$$

Also, note that $M(\mathcal{C}+\{i\}) = M(\mathcal{C})$ for $i \in M(\mathcal{C})$. In addition, it holds that $M(\mathcal{C}+\{i\}) = M(\mathcal{C}) \cup \{i\}$ and $|M(\mathcal{C}+\{i\})| = |M(\mathcal{C})| + 1$ for $i \notin M(\mathcal{C})$. From (5.9) we have

$$\sum_{i=1}^d \Pi^{(\ell)}_{\mathcal{C}_1+\{i\},\mathcal{C}_2} = \overbrace{\sum_{i\in M(\mathcal{C})} \frac{\eta\big[M(\mathcal{C}),\overline{\mathbf{X}}\big]}{\eta(\varnothing,\overline{\mathbf{X}})} \cdot \frac{s^*!/[s^* - |M(\mathcal{C})|]!}{(2\ell)!/[2\ell - |M(\mathcal{C})|]!}}^{(i)} \\ + \underbrace{\sum_{i\notin M(\mathcal{C})} \frac{\eta\big[M(\mathcal{C}) \cup \{i\},\overline{\mathbf{X}}\big]}{\eta(\varnothing,\overline{\mathbf{X}})} \cdot \frac{s^*!/[s^* - |M(\mathcal{C})| - 1]!}{(2\ell)!/[2\ell - |M(\mathcal{C})| - 1]!}}_{(ii)}. \tag{5.10}$$

Now we characterize the relationship between $\eta\big[M(\mathcal{C}),\overline{\mathbf{X}}\big]$ and $\eta\big[M(\mathcal{C}) \cup \{i\},\overline{\mathbf{X}}\big]$ with $i \notin M(\mathcal{C})$. Let $\mathcal{S}_1, \mathcal{S}_2, \ldots, \mathcal{S}_{\eta[M(\mathcal{C}),\overline{\mathbf{X}}]} \subseteq \{1,\ldots,d\}$ be the distinct sets satisfying $|\mathcal{S}_j| = 2\ell - |M(\mathcal{C})|$, $M(\mathcal{C}) \cap \mathcal{S}_j = \varnothing$ and $\text{sign}\big(\overline{\mathbf{X}}_{\mathcal{S}_j \cup M(\mathcal{C}), \mathcal{S}_j \cup M(\mathcal{C})}\big) = \mathbf{1}_{2\ell \times 2\ell}$ for all $j \in \{1,\ldots,\eta\big[M(\mathcal{C}),\overline{\mathbf{X}}\big]\}$. By setting $\mathcal{S}^{\sharp} = \cup_{j=1}^{\eta[M(\mathcal{C}),\overline{\mathbf{X}}]} \mathcal{S}_j$, we have that

$$\sum_{i\notin M(\mathcal{C})} \eta\big[M(\mathcal{C}) \cup \{i\},\overline{\mathbf{X}}\big] = \sum_{i\in \mathcal{S}^{\sharp}} \eta\big[M(\mathcal{C}) \cup \{i\},\overline{\mathbf{X}}\big] = \sum_{i\in \mathcal{S}^{\sharp}} \sum_{j=1}^{\eta[M(\mathcal{C}),\overline{\mathbf{X}}]} \mathbb{1}(i \in \mathcal{S}_j)$$

$$= \sum_{j=1}^{\eta[M(\mathcal{C}),\overline{\mathbf{X}}]} \sum_{i\in \mathcal{S}^{\sharp}} \mathbb{1}(i \in \mathcal{S}_j) = \sum_{j=1}^{\eta[M(\mathcal{C}),\overline{\mathbf{X}}]} |\mathcal{S}_j| = \eta\big[M(\mathcal{C}),\overline{\mathbf{X}}\big] \cdot [2\ell - |M(\mathcal{C})|].$$

Here the first equality is from $\eta\big[M(\mathcal{C}) \cup \{i\},\overline{\mathbf{X}}\big] = 0$ for $i \notin \mathcal{S}^{\sharp}$, since in this case

$$\text{sign}\big(\overline{\mathbf{X}}_{M(\mathcal{C}) \cup \{i\}, M(\mathcal{C}) \cup \{i\}}\big) \neq \mathbf{1}_{|M(\mathcal{C})\cup\{i\}|,|M(\mathcal{C})\cup\{i\}|}.$$

The second equality holds because to calculate $\eta\big[M(\mathcal{C}) \cup \{i\},\overline{\mathbf{X}}\big]$, we only need to count the number of $\mathcal{S}_j$'s that include $i$. The last equality is from $|\mathcal{S}_j| = 2\ell - |M(\mathcal{C})|$. Therefore, for term (ii) in (5.10)

we have

$$\sum_{i \notin M(\mathcal{C})} \frac{\eta[M(\mathcal{C}) \cup \{i\}, \overline{\mathbf{X}}]}{\eta(\varnothing, \overline{\mathbf{X}})} \cdot \frac{s^*!/[s^* - |M(\mathcal{C})| - 1]!}{(2\ell)!/[2\ell - |M(\mathcal{C})| - 1]!} \quad (5.11)$$

$$= \frac{\eta[M(\mathcal{C}), \overline{\mathbf{X}}]}{\eta(\varnothing, \overline{\mathbf{X}})} \cdot (2\ell - |M(\mathcal{C})|) \cdot \frac{s^*!/[s^* - |M(\mathcal{C})| - 1]!}{(2\ell)!/[2\ell - |M(\mathcal{C})| - 1]!} = \frac{\eta[M(\mathcal{C}), \overline{\mathbf{X}}]}{\eta(\varnothing, \overline{\mathbf{X}})} \cdot \frac{s^*!/[s^* - |M(\mathcal{C})| - 1]!}{(2\ell)!/[2\ell - |M(\mathcal{C})|]!}.$$

Meanwhile, for term (i) in (5.10) we have

$$\sum_{i \in M(\mathcal{C})} \frac{\eta[M(\mathcal{C}), \overline{\mathbf{X}}]}{\eta(\varnothing, \overline{\mathbf{X}})} \cdot \frac{s^*!/[s^* - |M(\mathcal{C})|]!}{(2\ell)!/[2\ell - |M(\mathcal{C})|]!} = |M(\mathcal{C})| \cdot \frac{\eta[M(\mathcal{C}), \overline{\mathbf{X}}]}{\eta(\varnothing, \overline{\mathbf{X}})} \cdot \frac{s^*!/[s^* - |M(\mathcal{C})|]!}{(2\ell)!/[2\ell - |M(\mathcal{C})|]!}$$

$$= (|M(\mathcal{C})| - s^*) \cdot \frac{\eta[M(\mathcal{C}), \overline{\mathbf{X}}]}{\eta(\varnothing, \overline{\mathbf{X}})} \cdot \frac{s^*!/[s^* - |M(\mathcal{C})|]!}{(2\ell)!/[2\ell - |M(\mathcal{C})|]!} + s^* \cdot \frac{\eta[M(\mathcal{C}), \overline{\mathbf{X}}]}{\eta(\varnothing, \overline{\mathbf{X}})} \cdot \frac{s^*!/[s^* - |M(\mathcal{C})|]!}{(2\ell)!/[2\ell - |M(\mathcal{C})|]!}$$

$$= -\frac{\eta[M(\mathcal{C}), \overline{\mathbf{X}}]}{\eta(\varnothing, \overline{\mathbf{X}})} \cdot \frac{s^*!/[s^* - |M(\mathcal{C})| - 1]!}{(2\ell)!/[2\ell - |M(\mathcal{C})|]!} + s^* \cdot \frac{\eta[M(\mathcal{C}), \overline{\mathbf{X}}]}{\eta(\varnothing, \overline{\mathbf{X}})} \cdot \frac{s^*!/[s^* - |M(\mathcal{C})|]!}{(2\ell)!/[2\ell - |M(\mathcal{C})|]!}. \quad (5.12)$$

Plugging (5.11) and (5.12) into (5.10), we obtain

$$\sum_{i=1}^{d} \Pi^{(\ell)}_{\mathcal{C}_1 + \{i\}, \mathcal{C}_2} = s^* \cdot \frac{\eta[M(\mathcal{C}), \overline{\mathbf{X}}]}{\eta(\varnothing, \overline{\mathbf{X}})} \cdot \frac{s^*!/[s^* - |M(\mathcal{C})|]!}{(2\ell)!/[2\ell - |M(\mathcal{C})|]!} = s^* \Pi^{(\ell)}_{\mathcal{C}_1, \mathcal{C}_2}.$$

Thus, we conclude that $\mathbf{\Pi}^{(\ell)}$ satisfies all the constraints of the $\ell$-th level SoS program in (3.6) except $\mathbf{\Pi}^{(\ell)} \succeq \mathbf{0}$. We defer the verification of this constraint to the end of the proof. Next we calculate the value of objective function corresponding to $\mathbf{\Pi}^{(\ell)}$. Note that

$$\sum_{i,j=1}^{d} \overline{X}_{i,j} \cdot \Pi^{(\ell)}_{\{i\}, \{j\}} = \sum_{i,j=1}^{d} \overline{X}_{i,j} \cdot \mathrm{sign}(\overline{X}_{i,j}) \cdot \Pi^{(\ell)}_{\{i\}, \{j\}} = \sum_{i,j=1}^{d} \overline{X}_{i,j} \cdot \mathbb{1}(\overline{X}_{i,j} > 0) \cdot \Pi^{(\ell)}_{\{i\}, \{j\}},$$

where the first equality holds because by the definition of $\eta(\cdot, \cdot)$, it holds $\eta(\{i,j\}, \overline{\mathbf{X}}) = 0$ for $\overline{X}_{i,j} \leq 0$, which implies $\Pi^{(\ell)}_{\{i\}, \{j\}} = 0$ correspondingly. Moreover, we have

$$\sum_{i,j=1}^{d} \Pi^{(\ell)}_{\{i\}, \{j\}} = \sum_{j=1}^{d} \sum_{i=1}^{d} \Pi^{(\ell)}_{\{i\}, \{j\}} = \sum_{j=1}^{d} s^* \Pi^{(\ell)}_{\varnothing, \{j\}} = s^* \sum_{j=1}^{d} \Pi^{(\ell)}_{\{j\}, \varnothing} = s^* \cdot s^* \Pi^{(\ell)}_{\varnothing, \varnothing} = (s^*)^2,$$

where the third and second last equalities are from the constraint $\sum_{i=1}^{d} \Pi^{(\ell)}_{\mathcal{C}_1 + \{i\}, \mathcal{C}_2} = s^* \Pi^{(\ell)}_{\mathcal{C}_1, \mathcal{C}_2}$, while the last is from $\Pi^{(\ell)}_{\varnothing, \varnothing} = 1$. Similarly, we have

$$\sum_{i=1}^{d} \Pi^{(\ell)}_{\{i\}, \{i\}} = \sum_{i=1}^{d} \Pi^{(\ell)}_{\{i\}, \varnothing} = s^* \Pi^{(\ell)}_{\varnothing, \varnothing} = s^*,$$

where the first equality follows from the constraints $\Pi^{(\ell)}_{\mathcal{C}_1 + \{i, i\}, \mathcal{C}_2} = \Pi^{(\ell)}_{\mathcal{C}_1 + \{i\}, \mathcal{C}_2}$ and $\Pi^{(\ell)}_{\mathcal{C}_1, \mathcal{C}_2} = \Pi^{(\ell)}_{\mathcal{C}'_1, \mathcal{C}'_2}$ for $\mathcal{C}_1 + \mathcal{C}_2 = \mathcal{C}'_1 + \mathcal{C}'_2$, and the second is from $\sum_{i=1}^{d} \Pi^{(\ell)}_{\mathcal{C}_1 + \{i\}, \mathcal{C}_2} = s^* \Pi^{(\ell)}_{\mathcal{C}_1, \mathcal{C}_2}$. Recall that $|X_{i,j}| \geq \nu$ almost surely and the objective function is equivalent to

$$\frac{1}{s^*(s^*-1)} \sum_{i,j=1}^{d} X_{i,j} \Pi^{(\ell)}_{\{i\},\{j\}} = \frac{1}{s^*(s^*-1)} \sum_{i,j=1}^{d} \overline{X}_{i,j} \cdot \mathbb{1}\left(\overline{X}_{i,j} > 0\right) \cdot \Pi^{(\ell)}_{\{i\},\{j\}} - \frac{\nu}{s^*(s^*-1)} \sum_{i=1}^{d} \Pi^{(\ell)}_{\{i\},\{i\}}$$

$$\geq \frac{\nu}{s^*(s^*-1)} \sum_{i,j=1}^{d} \Pi^{(\ell)}_{\{i\},\{j\}} - \frac{\nu}{s^*(s^*-1)} \sum_{i=1}^{d} \Pi^{(\ell)}_{\{i\},\{i\}} = \frac{\nu[(s^*)^2 - s^*]}{s^*(s^*-1)} \geq \nu.$$

Hence, the objective value corresponding to $\mathbf{\Pi}^{(\ell)}$ is $\nu$. Because $\widehat{\beta} \in \mathcal{H}^{(\ell)}$ is the maximum of the $\ell$-th level SoS program or its relaxed versions, so far we obtain

$$\mathbb{P}\bigl(\widehat{\beta} \geq \nu \mid \mathbf{\Pi}^{(\ell)} \succeq 0\bigr) = 1. \tag{5.13}$$

In the sequel, we verify that $\mathbf{\Pi}^{(\ell)} \succeq 0$ holds with high probability. We invoke Lemma 6.3 of Meka et al. (2015), which considers a matrix $\mathbf{M}^{(\ell)} \in \mathbb{R}^{\sum_{j=0}^{\ell}\binom{d}{j} \times \sum_{j=0}^{\ell}\binom{d}{j}}$ indexed by sets $\mathcal{S}_1, \mathcal{S}_2 \subseteq \{1, \ldots, d\}$, which satisfies $M^{(\ell)}_{\mathcal{S}_1, \mathcal{S}_2} = \Pi^{(\ell)}_{\mathcal{C}_1, \mathcal{C}_2}$ for $\mathcal{S}_1 = M(\mathcal{C}_1)$ and $\mathcal{S}_2 = M(\mathcal{C}_2)$. Their result implies that under the distribution within $\mathcal{P}(s^*, d)$ specified at the beginning of our proof, $\mathbf{M}^{(\ell)} \succeq \mathbf{0}$ holds with probability at least $1/2$ for sufficiently large $s^*$ and $d$, and $s^* = o\bigl\{[d/(\log d)^2]^{1/2\ell}\bigr\}$. Note $\mathbf{M}^{(\ell)}$ is a submatrix of $\mathbf{\Pi}^{(\ell)}$, i.e.,

$$\mathbf{M}^{(\ell)} = \mathbf{\Pi}^{(\ell)}_{\{\mathcal{C}:|\mathcal{C}|=|M(\mathcal{C})|\}, \{\mathcal{C}:|\mathcal{C}|=|M(\mathcal{C})|\}}.$$

In other words, we can simultaneously permute the rows and columns of $\mathbf{\Pi}^{(\ell)}$, which are indexed by the collection $\mathcal{C}$'s that satisfy $|\mathcal{C}| = |M(\mathcal{C})|$, to the upper-left corner of $\mathbf{\Pi}^{(\ell)}$. Then $\mathbf{M}^{(\ell)}$ is identical to such a $\sum_{j=0}^{\ell}\binom{d}{j} \times \sum_{j=0}^{\ell}\binom{d}{j}$ upper-left submatrix of $\mathbf{\Pi}^{(\ell)}$. Meanwhile, note that by (5.8) we have

$$\mathbf{\Pi}^{(\ell)}_{\mathcal{C}_1,*} = \mathbf{\Pi}^{(\ell)}_{\mathcal{C}_2,*}, \ \mathbf{\Pi}^{(\ell)}_{*,\mathcal{C}_1} = \mathbf{\Pi}^{(\ell)}_{*,\mathcal{C}_2}, \quad \text{for all } |\mathcal{C}_1| = |M(\mathcal{C}_1)|, \ M(\mathcal{C}_1) = M(\mathcal{C}_2).$$

Here $\mathbf{\Pi}^{(\ell)}_{\mathcal{C},*}$ and $\mathbf{\Pi}^{(\ell)}_{*,\mathcal{C}}$ denote the row and column corresponding to collection $\mathcal{C}$. Thus, for any vector $\mathbf{u} \in \mathbb{R}^{d^{(\ell)}}$, we have

$$\mathbf{u}^\top \mathbf{\Pi}^{(\ell)} \mathbf{u} = \mathbf{u}^\top \left[ \sum_{\mathcal{C}_1:|\mathcal{C}_1|=|M(\mathcal{C}_1)|} \left( \sum_{\mathcal{C}'_1:M(\mathcal{C}'_1)=M(\mathcal{C}_1)} u_{\mathcal{C}'_1} \right) \mathbf{\Pi}^{(\ell)}_{*,\mathcal{C}_1} \right]$$

$$= \sum_{\mathcal{C}_2} u_{\mathcal{C}_2} \left[ \sum_{\mathcal{C}_1:|\mathcal{C}_1|=|M(\mathcal{C}_1)|} \left( \sum_{\mathcal{C}'_1:M(\mathcal{C}'_1)=M(\mathcal{C}_1)} u_{\mathcal{C}'_1} \right) \mathbf{\Pi}^{(\ell)}_{\mathcal{C}_2,\mathcal{C}_1} \right]$$

$$= \sum_{\mathcal{C}_2:|\mathcal{C}_2|=|M(\mathcal{C}_2)|} \left( \sum_{\mathcal{C}'_2:M(\mathcal{C}'_2)=M(\mathcal{C}_2)} u_{\mathcal{C}'_2} \right) \left[ \sum_{\mathcal{C}_1:|\mathcal{C}_1|=|M(\mathcal{C}_1)|} \left( \sum_{\mathcal{C}'_1:M(\mathcal{C}'_1)=M(\mathcal{C}_1)} u_{\mathcal{C}'_1} \right) \mathbf{\Pi}^{(\ell)}_{\mathcal{C}_2,\mathcal{C}_1} \right]$$

$$= \overline{\mathbf{u}}^\top \mathbf{M}^{(\ell)} \overline{\mathbf{u}}, \tag{5.14}$$

where $\overline{\mathbf{u}} \in \mathbb{R}^{\sum_{j=0}^{\ell}\binom{d}{j}}$ is indexed by sets and $\overline{u}_\mathcal{S} = \sum_{\mathcal{C}:M(\mathcal{C})=\mathcal{S}} u_\mathcal{C}$. Thus, from (5.14) and the fact that $\mathbf{M}^{(\ell)} \succeq \mathbf{0}$ with probability at least $1/2$, we have $\mathbf{\Pi}^{(\ell)} \succeq \mathbf{0}$ holds with the same probability. Moreover, according to (5.13) and our setting that $\beta^* = 0$, by Markov's inequality we have

$$\mathbb{E}|\widehat{\beta} - \beta^*| \geq \nu \cdot \mathbb{P}\bigl(\widehat{\beta} \geq \nu\bigr) \geq \nu \cdot \mathbb{P}\bigl(\widehat{\beta} \geq \nu \mid \mathbf{\Pi}^{(\ell)} \succeq \mathbf{0}\bigr) \cdot \mathbb{P}\bigl(\mathbf{\Pi}^{(\ell)} \succeq \mathbf{0}\bigr) \geq 1/2 \cdot \nu \tag{5.15}$$

for all $\widehat{\beta} \in \mathcal{H}^{(\ell)}$ and $s^* = o\{[d/(\log d)^2]^{1/2\ell}\}$. Recall $\nu$ is a positive constant and our construction of distributions are within $\mathcal{P}(s^*, d)$. Hence, we conclude the proof. $\square$

Finally, we prove Proposition 4.4.

*Proof of Proposition 4.4.* We have

$$\mathbb{P}(|\widehat{\beta}^{\max} - \beta^*| \geq t) \leq \mathbb{P}\Big(\Big|\sup_{i,j \in \{1,\ldots,d\}} X_{i,j} - \sup_{i,j \in \{1,\ldots,d\}} \Theta_{i,j}\Big| \geq t\Big)$$
$$\leq \mathbb{P}\Big(\sup_{i,j \in \{1,\ldots,d\}} |X_{i,j} - \Theta_{i,j}| \geq t\Big) \leq d^2 \cdot \mathbb{P}(|X_{i,j} - \Theta_{i,j}| \geq t), \quad (5.16)$$

where the last inequality follows from union bound. Since $\mathbb{E}X_{i,j} = \Theta_{i,j}$, we have $\mathbb{E}(X_{i,j} - \Theta_{i,j}) = 0$. Moreover, we know that $\|X_{i,j} - \Theta_{i,j}\|_{\psi_2} \leq 1$. By the definition of sub-Gaussian random variable, we have

$$\mathbb{P}(|X_{i,j} - \Theta_{i,j}| \geq t) \leq \exp(1 - Ct^2). \quad (5.17)$$

Substituting (5.17) into (5.16), we obtain

$$\mathbb{P}(|\widehat{\beta}^{\max} - \beta^*| \geq t) \leq d^2 \exp(1 - Ct^2) = \exp(1 - Ct^2 + 2\log d). \quad (5.18)$$

Setting the right hand side of (5.18) to be $1/d$, and solving for $t$, we obtain with probability at least $1 - 1/d$ that

$$|\widehat{\beta}^{\max} - \beta^*| \leq C\sqrt{\log d}.$$

This completes the proof. $\square$

## 5.2 Proof for Stochastic Block Model

In this section, we present the detailed proofs of the main results for edge probability estimation in stochastic block model. We need the following lemma from Arias-Castro and Verzelen (2014), which provides the sufficient conditions under which the hypotheses $H_0 : \beta_0^* = p_0$ and $H_1 : \beta_1^* = p_1$ are not distinguishable. Recall $\mathbf{A}$ denotes the adjacency matrix and $\mathcal{P}(s^*, d)$ denotes the distribution family specified in §2.2.

**Lemma 5.2.** We consider testing $H_0 : \beta_0^* = p_0$ against $H_1 : \beta_1^* = p_1$. For any test $\phi : \mathbb{R}^{d \times d} \to \{0, 1\}$ based on $\mathbf{A}$, assuming $(s^*)^2(p_1 - p_0)/(\sqrt{p_0}d) = o(1)$, $\limsup (p_1 - p_0)^2 s^*/[4p_0(1 - p_0)\log(d/s^*)] < 1$ and $\log(d/s^*)/(s^*p_0) = o(1)$, we have

$$\inf_{\phi} \max\{\mathbb{P}_0(\phi = 1), \mathbb{P}_1(\phi = 0)\} \geq 1/4,$$

where $\mathbb{P}_0, \mathbb{P}_1 \in \mathcal{P}(s^*, d)$ are distributions corresponding to $H_0$ and $H_1$.

Now we are ready to lay out the proof of Theorem 4.5.

*Proof of Theorem 4.5.* The proof strategy is similar to Theorem 4.1. In the sequel, we assume $\widetilde{\beta}^*$ is known, since the obtained lower bound implies the lower bound for unknown $\widetilde{\beta}^*$. We invoke Lemma 5.2 with $p_0 = \widetilde{\beta}^*$ and $p_1 = \widetilde{\beta}^* + \overline{\beta}$, where

$$\overline{\beta} = C\sqrt{1/s^* \cdot \log(d/s^*)}. \tag{5.19}$$

Then we have that for any test $\phi : \mathbb{R}^{d \times d} \to \{0, 1\}$ based on the adjacency matrix $\mathbf{A}$, it holds that

$$\inf_{\phi} \max\{\mathbb{P}_0(\phi = 1), \mathbb{P}_1(\phi = 0)\} \geq 1/4, \quad \text{for } (s^*)^2 \overline{\beta}/d = o(1) \text{ and } \log(d/s^*)/(s^* \widetilde{\beta}^*) = o(1). \tag{5.20}$$

It is easy to verify the conditions in (5.20) are implied by the conditions of Theorem 4.5 and (5.19). Following the derivation of (5.4) in the proof of Theorem 4.1, we consider a specific test $\overline{\phi}(\widehat{\beta})$ based on $\widehat{\beta}$, which is defined as $\overline{\phi}(\widehat{\beta}) = \mathbb{1}(\widehat{\beta} > \widetilde{\beta}^* + \overline{\beta}/2)$. We have

$$\inf_{\widehat{\beta}} \max\Big\{\mathbb{P}_0(|\widehat{\beta} - \beta_0^*| \geq \overline{\beta}/2), \ \mathbb{P}_1(|\widehat{\beta} - \beta_1^*| \geq \overline{\beta}/2)\Big\}$$

$$= \inf_{\widehat{\beta}} \max\Big\{\mathbb{P}_0(|\widehat{\beta} - \widetilde{\beta}^*| \geq \overline{\beta}/2), \ \mathbb{P}_1(|\widehat{\beta} - \widetilde{\beta}^* - \overline{\beta}| \geq \overline{\beta}/2)\Big\}$$

$$\geq \inf_{\widehat{\beta}} \max\Big\{\mathbb{P}_0[\overline{\phi}(\widehat{\beta}) = 1], \ \mathbb{P}_1[\overline{\phi}(\widehat{\beta}) = 0]\Big\} \geq \inf_{\phi} \max\{\mathbb{P}_0(\phi = 1), \mathbb{P}_1(\phi = 0)\} \geq 1/4, \tag{5.21}$$

where the equality is obtained by plugging $\beta_0^*$ and $\beta_1^*$. The first inequality holds becasuse $\overline{\phi}(\widehat{\beta}) = 1$ implies $|\widehat{\beta} - \widetilde{\beta}^*| \geq \overline{\beta}/2$, and $\overline{\phi}(\widehat{\beta}) = 0$ implies $|\widehat{\beta} - \widetilde{\beta}^* - \overline{\beta}| \geq \overline{\beta}/2$. From (5.21) we obtain

$$\inf_{\widehat{\beta}} \sup_{\mathbb{P} \in \mathcal{P}(s^*, d)} \mathbb{E}_{\mathbb{P}}|\widehat{\beta} - \beta^*| \geq \inf \max\Big\{\mathbb{E}_{\mathbb{P}_0}|\widehat{\beta} - \beta_0^*|, \ \mathbb{E}_{\mathbb{P}_1}|\widehat{\beta} - \beta_1^*|\Big\}$$

$$\geq \overline{\beta}/2 \cdot \inf_{\widehat{\beta}} \max\Big\{\mathbb{P}_0(|\widehat{\beta} - \beta_0^*| \geq \overline{\beta}/2), \ \mathbb{P}_1(|\widehat{\beta} - \beta_1^*| \geq \overline{\beta}/2)\Big\} \geq \overline{\beta}/8,$$

where $\overline{\beta}$ is defined in (5.19), the second inequality follows from Markov's inequality. This concludes the proof. □

In the following, we prove Proposition 4.6.

*Proof of Proposition 4.6.* The proof is similar to Proposition 4.2. We only need to note that $\mathbf{A} - \mathbb{E}[\mathbf{A}]$ is a symmetric matrix, whose entires within the upper-right triangle are independently sub-Gaussian and satisfy

$$\|A_{i,j} - \mathbb{E}A_{i,j}\|_{\psi_2} \leq 1, \quad \text{for all } i < j,$$

since $A_{i,j}$ is Bernoulli and $|A_{i,j} - \mathbb{E}A_{i,j}| \leq 1$. Then replacing $\mathbf{X}$ with $\mathbf{A}$ in the proof of Proposition 4.2, we reach the conclusion. □

In the following, we lay out the proof of Theorem 4.7.

*Proof of Theorem 4.7.* We consider a specific distribution in $\mathcal{P}(s^*, d)$ under which the edge probability $\beta^* = \widetilde{\beta}^* = 1/2$. Let $\overline{\mathbf{A}} = \mathbf{A} + \mathbf{I}_d$. Under such a distribution, we construct a matrix $\mathbf{\Pi}^{(\ell)} \in \mathbb{R}^{d^{(\ell)} \times d^{(\ell)}}$, which is a feasible solution to the $\ell$-th level SoS optimization problem with high probability. Then we prove the objective value corresponding to $\mathbf{\Pi}^{(\ell)}$ is one, which implies that the maximum of the respective SoS program is at least one with high probability.

Different from the proof of Theorem 4.3, we define the expansivity $\eta(\mathcal{S}, \overline{\mathbf{A}})$ of $\mathcal{S} \subseteq \{1, \ldots, d\}$ as the number of sets $\mathcal{S}' \subseteq \{1, \ldots, d\}$ satisfying $|\mathcal{S}'| = 2\ell$, $\mathcal{S} \subseteq \mathcal{S}'$ and $\overline{\mathbf{A}}_{\mathcal{S}', \mathcal{S}'} = \mathbf{1}_{2\ell, 2\ell}$. Note $\eta(\mathcal{S}, \overline{\mathbf{A}})$ is nonzero only if $\overline{\mathbf{A}}_{\mathcal{S}, \mathcal{S}} = \mathbf{1}_{|\mathcal{S}|, |\mathcal{S}|}$. Therefore, $\eta(\mathcal{S}, \overline{\mathbf{A}})$ gives the number of $\overline{\mathbf{A}}$'s submatrices which are extended from $\overline{\mathbf{A}}_{\mathcal{S}, \mathcal{S}}$ and have size $2\ell \times 2\ell$ with all entries being one. Recall that each entry $\Pi^{(\ell)}_{\mathcal{C}_1, \mathcal{C}_2}$ of $\mathbf{\Pi}^{(\ell)}$ are indexed by collections $\mathcal{C}_1$ and $\mathcal{C}_2$, and $M(\mathcal{C}_1)$ and $M(\mathcal{C}_1)$ are the corresponding sets, which have distinct elements. Similar to (5.8), we construct each entry of $\mathbf{\Pi}^{(\ell)}$ as

$$\Pi^{(\ell)}_{\mathcal{C}_1, \mathcal{C}_2} = \frac{\eta\big[M(\mathcal{C}_1) \cup M(\mathcal{C}_2), \overline{\mathbf{A}}\big]}{\eta(\varnothing, \overline{\mathbf{A}})} \cdot \frac{s^*!/[s^* - |M(\mathcal{C}_1) \cup M(\mathcal{C}_2)|]!}{(2\ell)!/[2\ell - |M(\mathcal{C}_1) \cup M(\mathcal{C}_2)|]!}. \tag{5.22}$$

Note that the construction of $\Pi^{(\ell)}_{\mathcal{C}_1, \mathcal{C}_2}$ is exactly the same as (5.8), except that we replace $\overline{\mathbf{X}}$ with $\overline{\mathbf{A}}$. Also, by the same calculation as in the proof of Theorem 4.3, we can verify $\mathbf{\Pi}^{(\ell)}$ defined in (5.22) satisfies the constraints of $\ell$-th level SoS optimization problem.

Next we calculate the value of objective function corresponding to $\mathbf{\Pi}^{(\ell)}$. Note that

$$\sum_{i,j=1}^d \overline{A}_{i,j} \Pi^{(\ell)}_{\{i\},\{j\}} = \sum_{i,j=1}^d \mathbb{1}(\overline{A}_{i,j} = 1) \cdot \Pi^{(\ell)}_{\{i\},\{j\}} = \sum_{i,j=1}^d \Pi^{(\ell)}_{\{i\},\{j\}}.$$

Here both equalities hold because according to the definition of $\eta(\cdot, \overline{\mathbf{A}})$, it holds that $\eta(\{i,j\}, \overline{\mathbf{A}}) = 0$ for $\overline{A}_{i,j} \neq 1$, which implies $\Pi^{(\ell)}_{\{i\},\{j\}} = 0$ correspondingly. Moreover, we have

$$\sum_{i,j=1}^d \Pi^{(\ell)}_{\{i\},\{j\}} = \sum_{j=1}^d \sum_{i=1}^d \Pi^{(\ell)}_{\{i\},\{j\}} = \sum_{j=1}^d s^* \Pi^{(\ell)}_{\varnothing,\{j\}} = s^* \sum_{j=1}^d \Pi^{(\ell)}_{\{j\},\varnothing} = s^* \cdot s^* \Pi^{(\ell)}_{\varnothing,\varnothing} = (s^*)^2,$$

where the third and second last equalities are from the constraint $\sum_{i=1}^d \Pi^{(\ell)}_{\mathcal{C}_1 + \{i\}, \mathcal{C}_2} = s^* \Pi^{(\ell)}_{\mathcal{C}_1, \mathcal{C}_2}$, while the last is from $\Pi^{(\ell)}_{\varnothing,\varnothing} = 1$. Recall that the objective function is equivalent to

$$\frac{1}{s^*(s^* - 1)} \sum_{i,j=1}^d A_{i,j} \Pi^{(\ell)}_{\{i\},\{j\}} = \frac{1}{s^*(s^* - 1)} \sum_{i,j=1}^d \overline{A}_{i,j} \Pi^{(\ell)}_{\{i\},\{j\}} - \frac{1}{s^*(s^* - 1)} \sum_{i=1}^d \Pi^{(\ell)}_{\{i\},\{i\}}$$

$$= \frac{(s^*)^2 - s^*}{s^*(s^* - 1)} = 1.$$

Here the last equality holds because we have

$$\sum_{i=1}^d \Pi^{(\ell)}_{\{i\},\{i\}} = \sum_{i=1}^d \Pi^{(\ell)}_{\{i\},\varnothing} = s^* \Pi^{(\ell)}_{\varnothing,\varnothing} = s^*,$$

where the first equality follows from the constraints $\Pi^{(\ell)}_{\mathcal{C}_1+\{i,i\},\mathcal{C}_2} = \Pi^{(\ell)}_{\mathcal{C}_1+\{i\},\mathcal{C}_2}$ and $\Pi^{(\ell)}_{\mathcal{C}_1,\mathcal{C}_2} = \Pi^{(\ell)}_{\mathcal{C}'_1,\mathcal{C}'_2}$ for $\mathcal{C}_1 + \mathcal{C}_2 = \mathcal{C}'_1 + \mathcal{C}'_2$, and the second is from $\sum_{i=1}^{d} \Pi^{(\ell)}_{\mathcal{C}_1+\{i\},\mathcal{C}_2} = s^* \Pi^{(\ell)}_{\mathcal{C}_1,\mathcal{C}_2}$. Therefore, the objective value corresponding to $\mathbf{\Pi}^{(\ell)}$ is one. Because $\widehat{\beta} \in \mathcal{H}^{(\ell)}$ is the maximum of the $\ell$-th level SoS program or its relaxed versions, so far we obtain

$$\mathbb{P}\big(\widehat{\beta} \geq 1 \mid \mathbf{\Pi}^{(\ell)} \succeq \mathbf{0}\big) = 1. \tag{5.23}$$

According to the same proof of Theorem 4.3, we have $\mathbf{\Pi}^{(\ell)} \succeq \mathbf{0}$ holds with probability at least $1/2$ for $s^* = o\big\{[d/(\log d)^2]^{1/2\ell}\big\}$. Also, according to (5.23) and our setting that $\beta^* = 1/2$, from Markov's inequality we have

$$\mathbb{E}\big|\widehat{\beta} - \beta^*\big| \geq 1/2 \cdot \mathbb{P}\big(|\widehat{\beta} - \beta^*| \geq 1/2\big) \geq 1/2 \cdot \mathbb{P}\big(\widehat{\beta} \geq 1 \mid \mathbf{\Pi}^{(\ell)} \succeq \mathbf{0}\big) \cdot \mathbb{P}\big(\mathbf{\Pi}^{(\ell)} \succeq \mathbf{0}\big) \geq 1/4, \tag{5.24}$$

for any $\widehat{\beta} \in \mathcal{H}^{(\ell)}$ and $s^* = o\big\{[d/(\log d)^2]^{1/2\ell}\big\}$. Recall that our construction of distribution is within $\mathcal{P}(s^*, d)$. Hence we conclude the proof. $\square$

# 6 Conclusions

In this paper, we investigate the statistical limits of convex relaxations for two statistical problems: mean estimation for sparse principal submatrix and edge probability estimation for stochastic block model. Different from existing works, which consider the statistical limits of general polynomial-time algorithms, we instead characterize the loss in statistical rates incurred by a broad family of convex relaxations. At the core of our main theoretical results is a construction-based proof, which does not hinge on any unproven hardness hypotheses. Our conclusion is that in order to attain computational tractability with convex relaxations, under particular regimes we have to compromise the statistical optimality.